# Induction of First-Order Decision Lists:
# Results on Learning the Past Tense of English Verbs

**Raymond J. Mooney**                    MOONEY@CS.UTEXAS.EDU
**Mary Elaine Califf**                    MECALIFF@CS.UTEXAS.EDU
*Department of Computer Sciences, University of Texas*
*Austin, TX 78712-1188*

## Abstract

This paper presents a method for inducing logic programs from examples that learns a new class of concepts called *first-order decision lists*, defined as ordered lists of clauses each ending in a cut. The method, called FOIDL, is based on FOIL (Quinlan, 1990) but employs intensional background knowledge and avoids the need for explicit negative examples. It is particularly useful for problems that involve rules with specific exceptions, such as learning the past-tense of English verbs, a task widely studied in the context of the symbolic/connectionist debate. FOIDL is able to learn concise, accurate programs for this problem from significantly fewer examples than previous methods (both connectionist and symbolic).

## 1. Introduction

Inductive logic programming (ILP) is a growing subtopic of machine learning that studies the induction of Prolog programs from examples in the presence of background knowledge (Muggleton, 1992; Lavrač & Džeroski, 1994). Due to the expressiveness of first-order logic, ILP methods can learn relational and recursive concepts that cannot be represented in the attribute/value representations assumed by most machine-learning algorithms. ILP methods have successfully induced small programs for sorting and list manipulation (Shapiro, 1983; Sammut & Banerji, 1986; Muggleton & Buntine, 1988; Quinlan & Cameron-Jones, 1993) as well as produced encouraging results on important applications such as predicting protein secondary structure (Muggleton, King, & Sternberg, 1992) and automating the construction of natural-language parsers (Zelle & Mooney, 1994b).

However, current ILP techniques make important assumptions that restrict their application. Below are three common assumptions:

1. Background knowledge is provided in *extensional* form as a set of ground literals.

2. Explicit negative examples of the target predicate are available.

3. The target program is expressed in "pure" Prolog where clause-order is irrelevant and procedural operators such as cut (!) are disallowed.

The currently most well-known and successful ILP systems, GOLEM (Muggleton & Feng, 1990) and FOIL (Quinlan, 1990), both make all three of these assumptions. However, each of these assumptions brings significant limitations since:

1. An adequate extensional representation of background knowledge is frequently infinite or intractably large.

2. Explicit negative examples are frequently unavailable and an adequate set of negative examples computed using a closed-world assumption is infinite or intractably large.

3. Concise representation of many concepts requires the use of clause-ordering and/or cuts (Bergadano, Gunetti, & Trinchero, 1993).

This paper presents a new ILP method called Foidl (First-Order Induction of Decision Lists) which helps overcome each of these limitations by incorporating the following properties:

1. Background knowledge is represented *intensionally* as a logic program.

2. No explicit negative examples need be supplied or constructed. An assumption of *output completeness* can be used instead to implicitly determine if a hypothesized clause is overly-general and, if so, to quantify the degree of over-generality by simply estimating the number of negative examples covered.

3. A learned program can be represented as a *first-order decision list*, an ordered set of clauses each ending with a cut. This representation is very useful for problems that are best represented as general rules with specific exceptions.

As its name implies, Foidl is closely related to Foil and follows a similar top-down, greedy specialization guided by an information-gain heuristic. However, the algorithm is substantially modified to address the three advantages listed above. The use of intensional background knowledge is fairly straightforward and has been incorporated in previous Foil derivatives (Lavrač & Džeroski, 1994; Pazzani & Kibler, 1992; Zelle & Mooney, 1994b),

The development of Foidl was motivated by a failure we observed when applying existing ILP methods to a particular problem, that of learning the past tense of English verbs. This problem has been studied fairly extensively using both connectionist and symbolic methods (Rumelhart & McClelland, 1986; MacWhinney & Leinbach, 1991; Ling, 1994); however, previous efforts used specially-designed feature-based encodings that impose a fixed limit on the length of words and fail to capture the position-independence of the underlying transformation. We believed that representing the problem as constructing a logic program for the predicate `past(X,Y)` where `X` and `Y` are words represented as lists of letters (e.g `past([a,c,t]`, `[a,c,t,e,d])`, `past([a,c,h,e]`, `[a,c,h,e,d])`, `past([a,r,i,s,e]`, `[a,r,o,s,e])`) would produce much better results. However, due to the limitations mentioned above, we were unable to get reasonable results from either Foil or Golem. However, by overcoming these limitations, Foidl is able to learn highly accurate programs for the past-tense problem from many fewer examples than required by previous methods.

The remainder of the paper is organized as follows. Section 2 provides important background material on Foil and on the past-tense learning problem. Section 3 presents the Foidl algorithm and details how it incorporates the three advantages discussed above. Section 4 presents our results on learning the past-tense of English verbs demonstrating that Foidl out-performs all previous methods on this problem. Section 5 reviews related work, Section 6 discusses limitations and future directions, and Section 7 summarizes and presents our conclusions.

## 2. Background

Since FOIDL is based on FOIL, this section presents a brief review of this important ILP system; Quinlan (1990), Quinlan and Cameron-Jones (1993), and Cameron-Jones and Quinlan (1994) provide a more complete description. The section also presents a brief review of previous work on the English past tense problem.

### 2.1 FOIL

FOIL learns a function-free, first-order, Horn-clause definition of a *target* predicate in terms of itself and other *background* predicates. The input consists of extensional definitions of these predicates as tuples of constants of specified types. For example, input appropriate for learning a definition of list membership is:

```
member(Elt,Lst): { <a,[a]>, <a,[a,b]>, <b,[a,b]>, <a,[a,b,c]>, ...}
components(Lst,Elt,Lst): { <[a],a,[]>, <[a,b],a,[b]>, <[a,b,c],a,[b,c]> ...}
```

where `Elt` is a type denoting possible elements which includes `a,b,c`, and `d`; `Lst` is a type defined as consisting of flat lists containing up to three of these elements; and `components(A,B,C)` is a background predicate which is true iff `A` is a list whose first element is `B` and whose rest is the list `C` (this must be provided in place of a function for list construction). FOIL also requires negative examples of the target concept, which can be supplied directly or computed using a closed-world assumption. For the example, the closed-world assumption would produce all pairs of the form `<Elt,Lst>` that are not explicitly provided as positive examples (e.g., `<b,[a]>`).

Given this input, FOIL learns a program one clause at a time using a greedy-covering algorithm that can be summarized as follows:

Let *positives-to-cover* = positive examples.
While *positives-to-cover* is not empty
      Find a clause, $C$, that covers a preferably large subset of *positives-to-cover*
          but covers no negative examples.
      Add $C$ to the developing definition.
      Remove examples covered by $C$ from *positives-to-cover*.

For example, a clause that might be learned for `member` during one iteration of this loop is:

```
member(A,B) :- components(B,A,C).
```

since it covers all positive examples where the element is the first one in the list but does not cover any negatives. A clause that could be learned to cover the remaining examples is:

```
member(A,B) :- components(B,C,D), member(A,D).
```

Together these two clauses constitute a correct program for `member`.

The "find a clause" step is implemented by a general-to-specific hill-climbing search that adds antecedents to the developing clause one at a time. At each step, it evaluates possible literals that might be added and selects one that maximizes an information-gain heuristic. The algorithm maintains a set of tuples that satisfy the current clause and includes bindings for any new variables introduced in the body. The following pseudocode summarizes the procedure:

Initialize $C$ to $R(V_1, V_2, ..., V_k)$ :-. where $R$ is the target predicate with arity $k$.
Initialize $T$ to contain the positive tuples in *positives-to-cover* and all the negative tuples.
While $T$ contains negative tuples
    Find the best literal $L$ to add to the clause.
    Form a new training set $T'$ containing for each tuple $t$ in $T$ that satisfies $L$,
        all tuples of the form $t \cdot b$ ($t$ and $b$ concatenated) where $b$ is a set of bindings
        for the new variables introduced by $L$ such that the literal is satisfied
        (i.e., matches a tuple in the extensional definition of its predicate).
    Replace $T$ by $T'$.

Foil considers adding literals for all possible variablizations of each predicate as long as type restrictions are satisfied and at least one of the arguments is an existing variable bound by the head or a previous literal in the body. Literals are evaluated based on the number of positive and negative tuples covered, preferring literals that cover many positives and few negatives. Let $T_+$ denote the number of positive tuples in the set $T$ and define:

$$I(T) = -\log_2(T_+/|T|). \tag{1}$$

The chosen literal is then the one that maximizes:

$$gain(L) = s \cdot (I(T) - I(T')), \tag{2}$$

where $s$ is the number of tuples in $T$ that have extensions in $T'$ (i.e., the number of current positive tuples covered by $L$).

Foil also includes many additional features such as: heuristics for pruning the space of literals searched, methods for including equality, negation as failure, and useful literals that do not immediately provide gain (*determinate literals*), pre-pruning and post-pruning of clauses to prevent over-fitting, and methods for ensuring that induced programs will terminate. The papers referenced above should be consulted for details on these and other features.

## 2.2 Learning the Past Tense of English Verbs

Rumelhart and McClelland (1986) were the first to build a computational model of past-tense learning using the classic perceptron algorithm and a special phonemic encoding of words employing so-called Wickelphones and Wickelfeatures. Their general goal was to show that connectionist models could account for interesting language-learning behavior that was previously thought to require explicit rules. This model was heavily criticized by opponents of the connectionist approach to language acquisition for the relatively poor results achieved and the heavily-engineered representations and training techniques employed (Pinker & Prince, 1988; Lachter & Bever, 1988). MacWhinney and Leinbach (1991) attempted to address some of these criticisms by using a standard multi-layer backpropagation learning algorithm and a simpler UNIBET encoding of phonemes (in which each of 36 phonemes is encoded as a single ASCII character).

Ling and Marinov (1993) and Ling (1994) criticize all of the current connectionist models of past-tense acquisition for heavily-engineered representations and poor experimental methodology. They present more systematic results on a system called SPA (Symbolic Pattern Associator) which uses a slightly modified version of C4.5 (Quinlan, 1993) to build a forest of decision trees that maps a fixed-length input pattern to a fixed-length output pattern. Ling's (1994) head-to-head results show that SPA generalizes significantly better than backpropagation on a number of variations of the problem employing different phonemic encodings (e.g., 76% vs. 56% given 500 training examples).

However, all of this previous work encodes the problem as fixed-length pattern association and fails to capture the generativity and position-independence of the true transformation. For example, they use 15-letter patterns like:

```
a,c,t,_,_,_,_,_,_,_,_,_,_,_,_ => a,c,t,e,d,_,_,_,_,_,_,_,_,_,_
```

or in UNIBET phonemic encoding:

```
&,k,t,_,_,_,_,_,_,_,_,_,_,_,_ => &,k,t,I,d,_,_,_,_,_,_,_,_,_,_
```

where a separate decision tree or output unit is used to predict each character in the output pattern from all of the input characters. Therefore, learning general rules, such as "add 'ed'," must be repeated at each position where a word can end, and words longer than 15 characters cannot be handled. Also, the best results with SPA exploit a highly-engineered feature template and a modified version of C4.5's default leaf-labeling strategy that tailor it to string transformation problems.

Although ILP methods seem more appropriate for this problem, our initial attempts to apply FOIL and GOLEM to past-tense learning gave very disappointing results (Califf, 1994). Below, we discuss how the three problems listed in the introduction contribute to the difficulty of applying current ILP methods to this problem.

In principle, a background predicate for `append` is sufficient for constructing accurate past-tense programs when incorporated with an ability to include constants as arguments or, equivalently, an ability to add literals that bind variables to specific constants (called *theory constants* in FOIL). However, a background predicate that does not allow appending with the empty list is more appropriate. We use a predicate called `split(A, B, C)` which splits a list A into two non-empty sublists B and C. An intensional definition for `split` is:

```
split([X, Y | Z], [X] , [Y | Z]).
split([X | Y], [X | W], Z) :- split(Y,W,Z).
```

Using `split`, an "add 'ed'" rule can be represented as:

```
past(A,B) :- split(B,A,[e,d]).
```

which, in FOIL, is learned in the form:

```
past(A,B) :- split(B,A,C), C = [e,d].
```

Providing an extensional definition of `split` that includes all possible strings of 15 or fewer characters (at least $10^{21}$ strings) is clearly intractable. However, providing a partial definition that includes all possible splits of strings that actually appear in the training corpus is possible and generally sufficient. Therefore, providing adequate extensional background knowledge is cumbersome and requires careful engineering; however, it is not the major problem.

Supplying an appropriate set of negative examples is more problematic. Using a closed-world assumption to produce all pairs of words in the training set where the second is not the past-tense of the first is feasible but not very useful. In this case, the clause:

```
past(A,B) :- split(B,A,C).
```

is very likely to be learned since it covers most of the positives but very few (if any) negatives since it is unlikely that a word is a prefix of another word which is not its past tense. However, this clause is useless for producing the past tense of novel verbs, and, in this domain, accuracy must be measured by the ability to actually generate correct output for novel inputs, rather than the ability to classify pre-supplied tuples of arguments as positive or negative. The obvious solution of supplying all other strings of 15 characters or less as negative examples of the past tense of each word is clearly intractable. Providing specially constructed "near-miss" negative examples such as `past([a,c,h,e],[a,c,h,e,e,d])`, is very helpful, but requires careful engineering that exploits detailed prior knowledge of the problem.

In order to address the problem of negative examples, when Quinlan (1994) applied FOIL to this problem, he employed a different target predicate for representing the past-tense transformation.[1] He used a three-place predicate `past(X,Y,Z)` which is true iff the input word X is transformed into past-tense form by removing its current ending Y and substituting the ending Z; for example: `past([a,c,t], [], [e,d])`, `past([a,r,i,s,e], [i,s,e], [o,s,e])`. A simple preprocessor can map data for the two-place predicate into this form. Since a sample of 500 verb pairs contains about 30-40 different end fragments, this results in a more manageable number of closed-world negatives, approximately 1000 for every positive example in the training set. Using this approach on UNIBET phonemic encodings, Quinlan obtained slightly better results than Ling's best SPA results that exploited a highly-engineered feature template (83.3% vs. 82.8% with 500 training examples) and significantly better than SPA's normal results (76.3%). Although the three-place target predicate incorporates some knowledge about the desired transformation, it arguably requires less representation engineering than most previous methods.

However, Quinlan (1994) notes that his results are still hampered by FOIL's inability to exploit clause order. For example, when using normal alphabetic encoding, FOIL quickly learns a clause sufficient for regular verbs:

```
past(A,B,C) :- B=[], C=[e,d].
```

However, since this clause still covers a fair number of negative examples due to many irregular verbs, it continues to add literals. As a result, FOIL creates a number of specialized versions of this clause that together still fail to capture the generality of the underlying default rule. This problem is compounded by FOIL's inability to add constraints such as "does not end in 'e'." Since FOIL separates the addition of literals containing variables and the binding of variables to constants using literals of the form $V = c$, it cannot learn clauses like:

```
past(A,B,C) :- B=[], C=[e,d], not(split(A,D,[e])).
```

Since a word can be split in several ways, this is clearly not equivalent to the learnable clause:

```
past(A,B,C) :- B=[], C=[e,d], not(split(A,D,E)), E /= [e].
```

---

1. Quinlan's work on this problem was motivated by our own early attempts to use FOIL.

Consequently, it must approximate the true rule by learning many clauses of the form:

```
past(A,B,C) :- B=[], C=[e,d], split(A,D,E), E = [b].
past(A,B,C) :- B=[], C=[e,d], split(A,D,E), E = [d].
...
```

As a result, FOIL generated overly-complex programs containing more than 40 clauses for both the phonemic and alphabetic versions of the problem.

However, an experienced Prolog programmer would exploit clause order and cuts to write a concise program that first handles the most-specific exceptions and falls through to more-general default rules if the exceptions fail to apply. For example, the program:

```
past(A,B) :- split(A,C,[e,e,p]), split(B,C,[e,p,t]), !.
past(A,B) :- split(A,C,[y]), split(B,C,[i,e,d]), !.
past(A,B) :- split(A,C,[e]), split(B,A,[d]), !.
past(A,B) :- split(B,A,[e,d]).
```

can be summarized as:

> If the word ends in "eep," then replace "eep" with "ept" (e.g., sleep, slept),
> else, if the word ends in "y," then replace "y" with "ied"
> else, if the word ends in "e," add "d"
> else, add "ed."

FOIDL can directly learn programs of this form, i.e., ordered sets of clauses each ending in a cut. We call such programs first-order decision lists due to the similarity to the propositional *decision lists* introduced by Rivest (1987). FOIDL uses the normal binary target predicate and requires no explicit negative examples. Therefore, we believe it requires *significantly* less representation engineering than all previous work in the area.

## 3. FOIDL Induction Algorithm

As stated in the introduction, FOIDL adds three major features to FOIL: 1) Intensional specification of background knowledge, 2) Output completeness as a substitute for explicit negative examples, and 3) Support for learning first-order decision lists. The following subsections describe the modifications made to incorporate these features.

### 3.1 Intensional Background

As described above, FOIL assumes background predicates are provided with extensional definitions; however, this is burdensome and frequently intractable. Providing an intensional definition in the form of general Prolog clauses is generally preferable. For example, instead of providing numerous tuples for the **components** predicates, it is easier to give the intensional definition:

```
components([A | B], A, B).
```

Intentional background definitions are not restricted to function-free pure Prolog and can exploit all features of the language.

Modifying Foil to use intensional background is straightforward. Instead of matching a literal against a set of tuples to determine whether or not it covers an example, the Prolog interpreter is used in an attempt to prove that the literal can be satisfied using the intensional definitions. Unlike Foil, expanded tuples are not maintained and positive and negative examples of the target concept are reproved for each alternative specialization of the developing clause. Therefore, the pseudocode for learning a clause is simply:

Initialize $C$ to $R(V_1, V_2, ..., V_k)$ :-. where $R$ is the target predicate with arity $k$.
Initialize $T$ to contain the examples in *positives-to-cover* and all the negative examples.
While $T$ contains negative tuples
      Find the best literal $L$ to add to the clause.
      Let $T'$ be the subset of examples in $T$ that can still be proved as instances of the
          target concept using the specialized clause.
      Replace $T$ by $T'$

Since expanded tuples are not produced, the information-gain heuristic for picking the best literal is simply:

$$gain(L) = |T'| \cdot (I(T) - I(T')). \tag{3}$$

## 3.2 Output Completeness and Implicit Negatives

In order to overcome the need for explicit negative examples, a mode declaration for the target concept must be provided (i.e., a specification whether each argument is an input (+) or an output (-)). An assumption of *output completeness* can then be made, indicating that for every unique input pattern in the training set, the training set includes all of the correct output patterns. Therefore, any other output which a program produces for a given input can be assumed to represent a negative example. This does not require that all positive examples be part of the training set, only that for each unique input pattern in the training set, all other positive examples with that input pattern (if any) must also be in the training set. This assumption is trivially met if the predicate represents a function with a single unique output for each input.

For example, an assumption of output completeness for the mode declaration `past(+,-)` indicates that all of the correct past-tense forms are included for each input word in the training set. For predicates representing functions, such as `past`, this implies that the output for each example is unique and that all other outputs implicitly represent negative examples. However, output completeness can also be applied to non-functional cases such as `append(-,-,+)`, indicating that all possible pairs of lists that can be appended together to produce a list are included in the training set (e.g., `append([],[a,b],[a,b])`, `append([a],[b],[a,b])`, `append([a,b],[],[a,b])`).

Given an output completeness assumption, determining if a clause is overly-general is straightforward. For each positive example, an *output query* is made to determine all outputs for the given input (e.g., `past([a,c,t], X)`). If any outputs are generated that are not positive examples, the clause still covers negative examples and requires further specialization. Note that intensional interpretation of learned clauses is required in order to answer output queries.

In addition, in order to compute the gain of alternative literals during specialization, the negative coverage of a clause needs to be quantified. Each incorrect answer to an output query which is ground (i.e., contains no variables) clearly counts as a single negative example (e.g., `past([a,c,h,e], [a,c,h,e,e,d])`). However, output queries will frequently produce answers with universally quantified variables. For example, given the overly-general clause `past(A,B) :- split(A,C,D).`, the query `past([a,c,t], X)` generates the answer `past([a,c,t], Y)`. This implicitly represents coverage of an infinite number of negative examples. In order to quantify negative coverage, FOIDL uses a parameter $u$ to represent a bound on the number of possible terms. Since the set of all possible terms (the Herbrand universe of the background knowledge together with the examples) is generally infinite, $u$ is meant to represent a heuristic estimate of the finite number of these terms that will ever actually occur in practice (e.g., the number of distinct words in English). The negative coverage represented by a non-ground answer to an output query is then estimated as $u^v - p$, where $v$ is the number of variable arguments in the answer and $p$ is the number of positive examples with which the answer unifies. The $u^v$ term stands for the number of unique ground outputs represented by the answer (e.g., the answer `append(X,Y,[a,b])` stands for $u^2$ different ground outputs) and the $p$ term stands for the number of these that represent positive examples. This allows FOIDL to quantify coverage of large numbers of implicit negative examples without ever explicitly constructing them. It is generally sufficient to estimate $u$ as a fairly large constant (e.g., 1000), and empirically the method is not very sensitive to its exact value as long as it is significantly greater than the number of ground outputs ever generated by a clause.

Unfortunately, this estimate is not sensitive enough. For example, both clauses

```
past(A,B) :- split(A,C,D).
past(A,B) :- split(B,A,C).
```

cover $u$ implicit negative examples for the output query `past([a,c,t], X)` since the first produces the answer `past([a,c,t], Y)` and the second produces the answer `past([a,c,t], [a,c,t | Y])`. However, the second clause is clearly better since it at least requires the output to be the input with some suffix added. Since there are presumably more words than there are words that start with "a-c-t" (assuming the total number of words is finite), the first clause should be considered to cover more negative examples. Therefore, arguments that are partially instantiated, such as `[a,c,t | Y]`, are counted as only a fraction of a variable when calculating $v$. Specifically, a partially instantiated output argument is scored as the fraction of its subterms that are variables, e.g., `[a,c,t | Y]` counts as only 1/4 of a variable argument. Therefore, the first clause above is scored as covering $u$ implicit negatives and the second as covering only $u^{1/4}$. Given reasonable values for $u$ and the number of positives covered by each clause, the literal `split(B,A,C)` will be preferred.

The revised specialization algorithm that incorporates implicit negatives is:

Initialize $C$ to $R(V_1, V_2, ..., V_k)$ :-. where $R$ is the target predicate with arity $k$.
Initialize $T$ to contain the examples in *positives-to-cover* and output queries for all
    positive examples.
While $T$ contains output queries
        Find the best literal $L$ to add to the clause.
        Let $T'$ be the subset of positive examples in $T$ that can still be proved as instances
            of the target concept using the specialized clause, plus the output queries in $T$

that still produce incorrect answers.
Replace $T$ by $T'$.

Literals are scored as described in the previous section except that $|T|$ is computed as the number of positive examples in $T$ plus the sum of the number of implicit negatives covered by each output query in $T$.

## 3.3 First-Order Decision Lists

As described above, first-order decision lists are ordered sets of clauses each ending in a cut. When answering an output query, the cuts simply eliminate all but the first answer produced when trying the clauses in order. Therefore, this representation is similar to propositional decision lists (Rivest, 1987), which are ordered lists of pairs (rules) of the form $(t_i, c_i)$ where the test $t_i$ is a conjunction of features and $c_i$ is a category label and an example is assigned to the category of the first pair whose test it satisfies.

In the original algorithm of Rivest (1987) and in CN2 (Clark & Niblett, 1989), rules are learned in the order they appear in the final decision list (i.e., new rules are appended to the end of the list as they are learned). However, Webb and Brkič (1993) argue for learning decision lists in the reverse order since most preference functions tend to learn more general rules first, and these are best positioned as default cases towards the end. They introduce an algorithm, *prepend*, that learns decision lists in reverse order and present results indicating that in most cases it learns simpler decision lists with superior predictive accuracy. FOIDL can be seen as generalizing *prepend* to the first-order case for target predicates representing functions. It learns an ordered sequence of clauses in reverse order, resulting in a program which produces only the first output generated by the first satisfied clause.

The basic operation of the algorithm is best illustrated by a concrete example. For alphabetic past-tense, the current algorithm easily learns the partial clause:

    past(A,B) :- split(B,A,C), C = [e,d].

However, as discussed in section 2.2, this clause still covers negative examples due to irregular verbs. However, it produces correct ground output for a subset of the examples (i.e., the regular verbs).[2] This is an indication that it is best to terminate this clause to handle these examples, and add earlier clauses in the decision list to handle the remaining examples. The fact that it produces incorrect answers for other output queries can be safely ignored in the decision-list framework since these can be handled by earlier clauses. Therefore, the examples correctly covered by this clause are removed from *positives-to-cover* and a new clause is begun. The literals that now provide the best gain are:

    past(A,B) :- split(B,A,C), C = [d].

since many of the irregulars are those that just add "d" (since they end in "e"). This clause also now produces correct ground output for a subset of the examples; however, it is not complete since it produces incorrect output for examples correctly covered by a previously learned clause (e.g., past([a,c,t], [a,c,t,d])). Therefore, specialization continues until all of these cases are also eliminated. This results in the clause:

---

2. Note that this is untrue until both of the literals are added to this initially empty clause.

```
past(A,B) :- split(B,A,C), C = [d], split(A,D,E), E = [e].
```

which is added to the front of the decision list and the examples it covers are removed from *positives-to-cover*. This approach ensures that every new clause produces correct outputs for some new subset of the examples but doesn't result in incorrect output for examples already correctly covered by previously learned clauses. This process continues adding clauses to the front of the decision list until all of the exceptions are handled and *positives-to-cover* is empty.

The resulting clause-specialization algorithm can now be summarized as follows:

Initialize $C$ to $R(V_1, V_2, ..., V_k)$ :-. where $R$ is the target predicate with arity $k$.
Initialize $T$ to contain the examples in *positives-to-cover* and output queries for all
    positive examples.
While $T$ contains output queries
    Find the best literal $L$ to add to the clause.
    Let $T'$ be the subset of positive examples in $T$ whose output query still produces
       a first answer that unifies with the correct answer, plus the output queries in $T$
       that either
           1) Produce a non-ground first answer that unifies with the correct answer, or
           2) Produce an incorrect answer but produce a correct answer using a
              previously learned clause.
    Replace $T$ by $T'$.

In many cases, this algorithm is able to learn accurate, compact, first-order decision lists for past tense, like the "expert" program shown in section 2.2. However, due to highly irregular verbs, the algorithm can encounter local-minima in which it is unable to find any literals that provide positive gain while still covering the required minimum number of examples.[3] This was originally handled by terminating search and memorizing any remaining uncovered examples as specific exceptions at the top of the decision list (e.g., past([a,r,i,s,e], [a,r,o,s,e]) :- !.). However, this can result in premature termination that prevents the algorithm from finding low-frequency regularities. For example, in the alphabetic version, the system can get stuck trying to learn the complex rule for when to double a final consonant (e.g., grab → grabbed) and fail to learn the rule for changing "y" to "ied" since this is actually less frequent.

The current version, like FOIL, tests if the learned clause meets a minimum-accuracy threshold; however, unlike FOIL, only counting as errors incorrect outputs for queries correctly answered by previously learned clauses. If it does not meet the threshold, the clause is thrown out and the positive examples it covers are memorized at the top of the decision list. The algorithm then continues to learn clauses for any remaining positive examples. This allows FOIDL to just memorize difficult irregularities, such as consonant doubling, and still continue on to learn other rules such as changing "y" to "ied."

If the minimum-accuracy threshold is met, the decision-list property is exploited in a final attempt to still learn a completely accurate program. If the negatives covered by the clause are all examples that were correctly covered by previously learned clauses, FOIDL

---

treats them as "exceptions to the exception to the rule" and returns them to *positives-to-cover* to be covered correctly again by subsequently learned clauses. For example, FOIDL frequently learns the clause:

```
past(A, B) :- split(A, C, [y]), split(B, C, [i, e, d]).
```

for changing "y" to "ied." However, this clause incorrectly covers a few examples that are correctly covered by the previously learned "add 'ed'" rule (e.g., bay → bayed; delay → delayed). Since these exceptions to the "y" to "ied" rule are a small percentage of the words that end in "y," the system keeps the rule and returns the examples that just add "ed" to *positives-to-cover*. Subsequently, rules such as:

```
past(A, B) :- split(B, A, [e, d]), split(A, D, [a, y]).
```

are learned to recover these examples, resulting in a program that is completely consistent with the training data. By setting the minimum clause-accuracy threshold to 50%, FOIDL only applies this *uncovering* technique when it results in covering more examples than it uncovers, thereby guaranteeing progress towards fitting all of the training examples.

## 3.4 Algorithmic and Implementation Details

This section briefly discusses a few additional details of the FOIDL algorithm and its implementation. This includes a discussion of the use of modes, types, weak literals, and theory constants. The current version of FOIL includes all of these features in basically the same form.

FOIDL makes use of types and modes to limit the space of literals searched. The argument of each predicate is typed and only literals whose previously-bound arguments are of the correct type are tested when specializing a clause. For example, split is given the types split(word,prefix,suffix), preventing the system from further splitting prefixes and suffixes and exploring arbitrary substrings of a word for regularities. Each predicate is also given a mode declaration, and only literals whose input arguments are all previously-bound variables are tested. For example, split is given the mode split(+,-,-), preventing a clause from creating new strings by appending together previously generated prefixes and suffixes.

In case no literal provides positive information gain, FOIDL gives a small bonus to literals that introduce new variables. However, the number of such *weak literals* that can be added in a row is limited by a user parameter (normally set to 1). For example, this allows the system to split a word into possible prefixes and suffixes, even though this may not provide gain until these substrings are constrained by subsequent literals.

Theory constants are provided for each type, and literals are tested for binding each existing variable to each constant of the appropriate type. For example, the literal X=[e,d] is generated if X is of type suffix. For our runs on past-tense, theory constants are included for every prefix and suffix that occurs in at least two words in the training data. This helps control training time by limiting the number of literals searched, but does not affect which literals are actually chosen since the minimum-clause-coverage test prevents FOIDL from choosing literals that don't cover at least two examples anyway.

Foidl is currently implemented in both Common Lisp and Quintus Prolog. Unlike the current Prolog version, the Common Lisp version supports learning recursive clauses[4] and output-completeness for non-functional target predicates. However, the Common Lisp version is significantly slower since it relies on an un-optimized Prolog interpreter and compiler written in Lisp (from Norvig, 1992). Consequently, all of the presented results are from the Prolog version running on a Sun SPARCstation 2.[5]

## 4. Experimental Results

To test Foidl's performance on the English past tense task, we ran experiments using the data which Ling (1994) made available in an appendix.

### 4.1 Experimental Design

The data used consist of 6939 English verb forms in both normal alphabetic form and UNIBET phoneme representation along with a label indicating the verb form (base, past tense, past participle, etc), a label indicating whether the form is regular or irregular, and the Francis-Kucera frequency of the verb. The data include 1390 distinct pairs of base and past tense verb forms. We ran three different experiments. In one we used the phonetic forms of all verbs. In the second we used the phonetic forms of the regular verbs only, because this is the easiest form of the task and because this is the only problem for which Ling provides learning curves. Finally, we ran trials using the alphabetic forms of all verbs. The training and testing followed the standard paradigm of splitting the data into testing and training sets and training on progressively larger samples of the training set. All results were averaged over 10 trials, and the testing set for each trial contained 500 verbs.

In order to better separate the contribution of using implicit negatives from the contribution of the decision list representation, we also ran experiments with IFoil, a variant of the system which uses intensional background and the output completeness assumption, but does not build decision lists.

We ran our own experiments with Foil, Foidl, and IFoil and compared those with the results from Ling. The Foil experiments were run using Quinlan's representation described in section 2.2. As in Quinlan (1994), negative examples were provided by using a randomly-selected 25% of those which could be generated using the closed world assumption.[6] All experiments with Foidl and IFoil used the standard default values for the various numeric parameters (term universe size, 1000; minimum clause coverage, 2; weak literal limit, 1). The differences among Foil, IFoil, and Foidl were tested for significance using a two-tailed paired t-test.

---

4. Handling intensional interpretation of recursive clauses for the target predicate requires some additional complexities that have not been discussed in this paper since they are not relevant to decision-lists, which are generally not recursive.

5. Both versions are available by anonymous FTP from net.cs.utexas.edu in the directory pub/mooney/foidl.

6. We replicated Quinlan's approach since memory limitations prevented us from using 100% of the generated negatives with larger training sets.

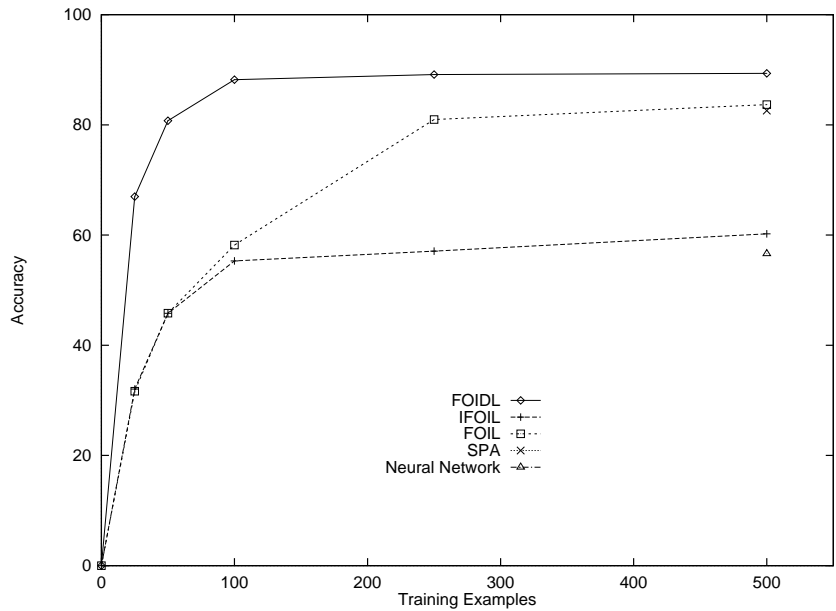

Figure 1: Accuracy on phonetic past tense task using all verbs

## 4.2 Results

The results for the phonetic task using both regular and irregular verbs are presented in Figure 1. The graph shows our results with FOIL, IFOIL, and FOIDL along with the best results from Ling, who did not provide a learning curve for this task. As expected, FOIDL out-performed the other systems on this task, surpassing Ling's best results with 500 examples with only 100 examples. IFOIL performed quite poorly, barely beating the neural network results despite effectively having 100% of the negatives as opposed to FOIL's 25%. This poor performance is due at least in part to overfitting the training data, because IFOIL lacks the noise-handling techniques of FOIL6. FOIL also has the advantage of the three-place predicate, which gives it a bias toward learning suffixes. IFOIL's poor performance on this task shows that the implicit negatives by themselves are not sufficient, and that some other bias such as decision lists or the three-place predicate and noise-handling is needed. The differences between FOIL and FOIDL are significant at the 0.01 level. Those between FOIDL and IFOIL are significant at the 0.001 level. The differences between FOIL and IFOIL are not significant with 100 training examples or less, but are significant at the 0.001 level with 250 and 500 examples.

Figure 2 presents accuracy results on the phonetic task using regulars only. The curves for SPA and the neural net are the results reported by Ling. Here again, FOIDL out-performed the other systems. This particular task demonstrated one of the problems with using closed-world negatives. In the regular past tense task, the second argument of Quinlan's 3-place predicate is always the same: an empty list. Therefore, if the constants are generated from the positive examples, FOIL will never produce rules which ground the second argument, since it cannot create negative examples with other constants in the second argument. This prevents the system from learning a rule to generate the past tense. In order

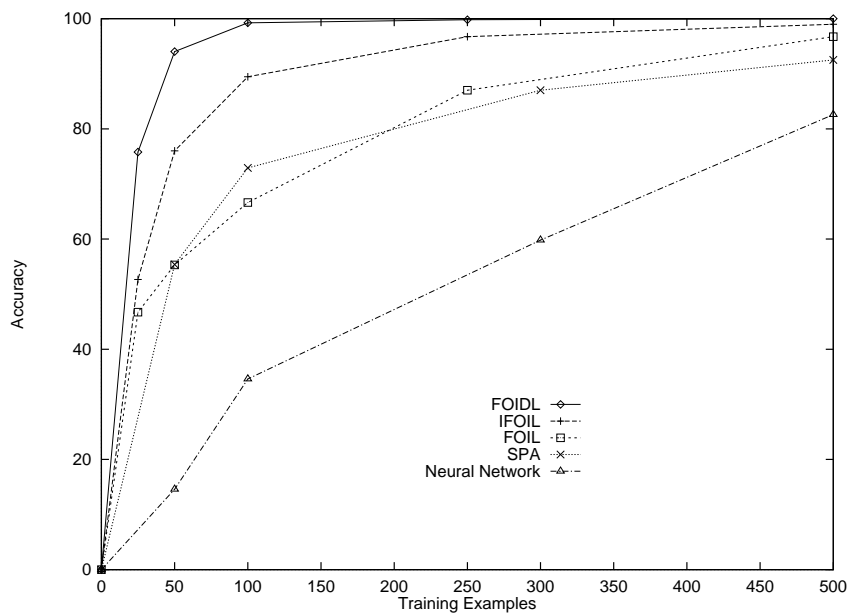

Figure 2: Accuracy on phonetic past tense task using regulars only to obtain the results reported here, we introduced extra constants for the second argument (specifically the constants for the third argument), enabling the closed world assumption to generate appropriate negatives. On this task, IFOIL does seem to gain some advantage over FOIL from being able to effectively use all of the negatives. The regularity of the data allows both IFOIL and FOIL to achieve over 90% accuracy at 500 examples. The differences between FOIL and FOIDL are significant at the 0.001 level, as are those between IFOIL and FOIDL. The differences between IFOIL and FOIL are not significant with 25 examples, and are significant at the 0.02 level with 500 examples, but are significant at the 0.001 level with 50-250 training examples.

Results for the alphabetic version appear in Figure 3. This is a task which has not typically been considered in the literature, but it is of interest to those concerned with incorporating morphology into natural language understanding systems which deal with text. It is also the most difficult task, primarily because of consonant doubling. Here we have results only for FOIDL, IFOIL, and FOIL. Because the alphabetic task is even more irregular that the full phonetic task, IFOIL again overfits the data and performs quite poorly. The differences between FOIL and FOIDL are significant at the 0.001 level with 25, 50, 250, and 500 examples, but only at the 0.1 level with 100 examples. The differences between IFOIL and FOIDL are all significant at the 0.001 level. Those between FOIL and IFOIL are not significant with 25 training examples and are significant only at the 0.01 level with 50 training examples, but are significant at the 0.001 level with 100 or more examples.

For all three of these tasks, FOIDL clearly outperforms the other systems, demonstrating that the first order decision list bias is a good one for this learning task. A sufficient set of negatives is necessary, and all five of these systems provide them in some way: the neural network and SPA both learn multiple-class classification tasks (which phoneme belongs in each position); FOIL uses the three-place predicate with closed world negatives; and IFOIL and FOIDL, of course, use the output completeness assumption. The primary importance of the implicit negatives is not that they provide an advantage over propositional and neural network systems, but that they enable first order systems to perform this task at all. Without them, some knowledge of the task is required. FOIDL's decision lists give it a significant added advantage, though this advantage is less apparent in the regular phonetic task, where there are no exceptions.

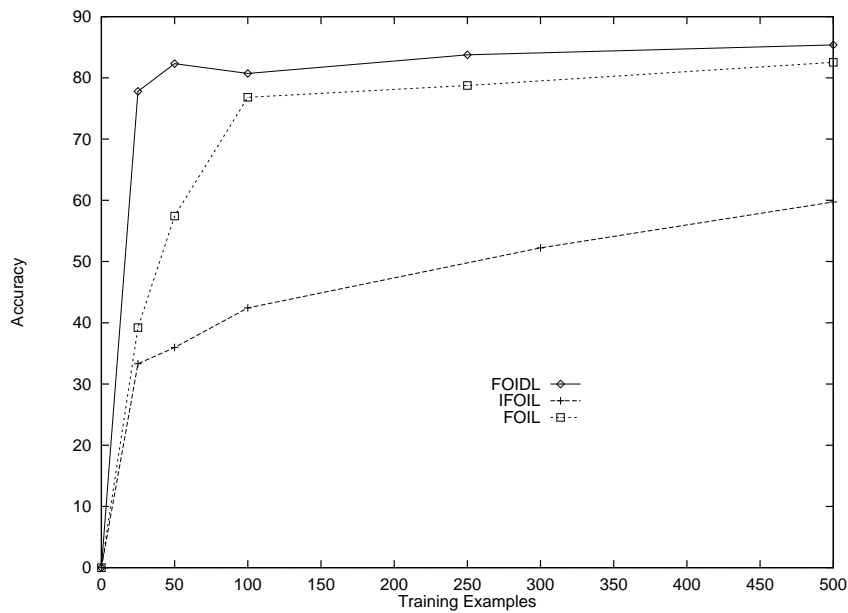

Figure 3: Accuracy on alphabetic past tense task

Clearly, FOIDL produces more accurate rules than the other systems, but another consideration is the complexity of the rule sets. For the ILP systems, two good measures of complexity are the number of rules and number of literals generated. Figure 4 shows the number of rules generated by FOIL, IFOIL, and FOIDL for the phonetic task using all verbs. The number of literals generated appears in Figure 5. Since we are interested in generalization and since FOIL does not attempt to fit all of the training data, these results do not include the rules FOIDL and IFOIL add in order to memorize individual exceptions.[7] Although the numbers are comparable with only a few examples, with increasing numbers of examples, the programs FOIL and IFOIL generate grow much faster than FOIDL's programs. The large number of rules/literals learned by IFOIL show its tendency to overfit the data.

FOIDL also generates very comprehensible programs. The following is an example program generated for the alphabetic version of the task using 250 examples (again excluding the memorized examples).

```
past(A,B) :- split(A,C,[e,p]), split(B,C,[p,t]),!.
past(A,B) :- split(A,C,[y]), split(B,C,[i,e,d]), split(A,D,[r,y]),!.
past(A,B) :- split(A,C,[y]), split(B,C,[i,e,d]), split(A,D,[l,y]),!.
```

---

7. Because of the large number of irregular pasts in English, FOIDL memorizes an average of 38 verbs per trial with 500 examples.

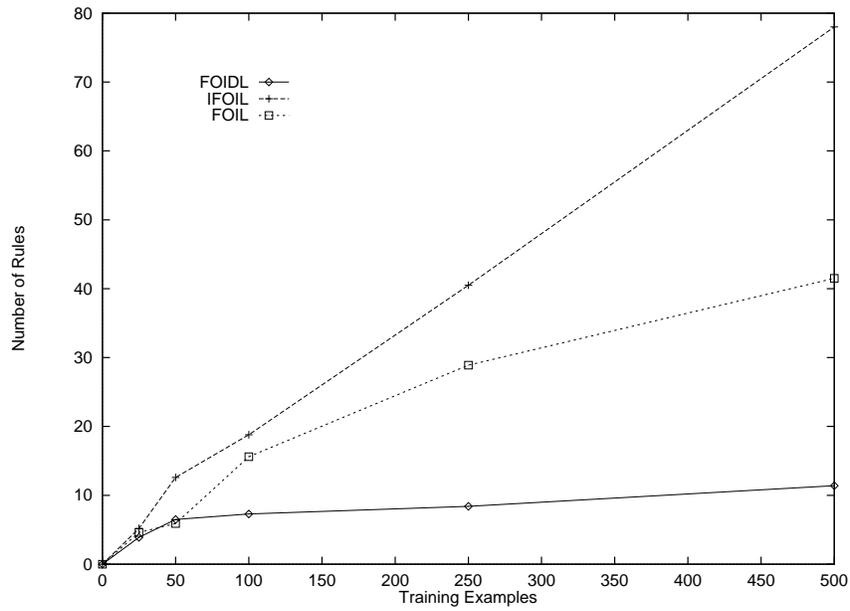

Figure 4: Number of rules created for phonetic past tense task

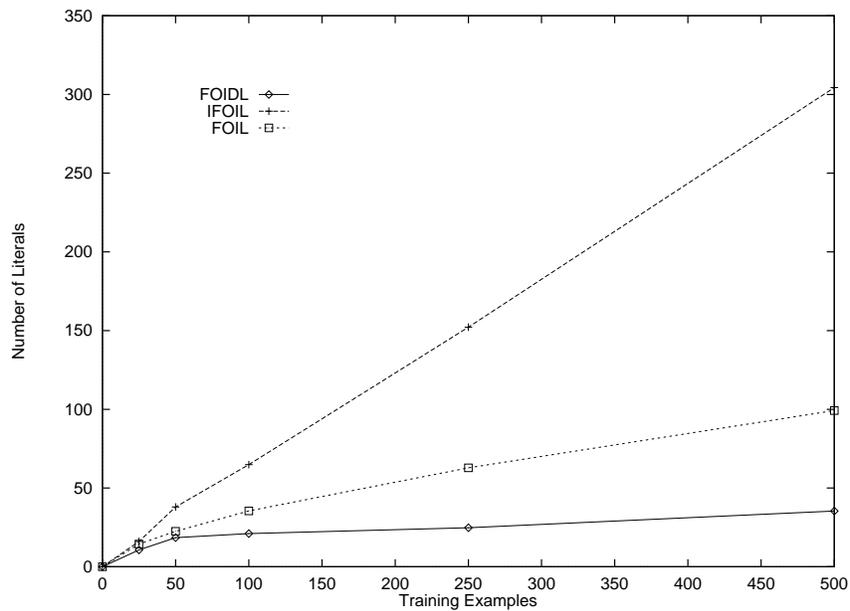

Figure 5: Number of literals created for phonetic past tense task

```
past(A,B) :- split(B,A,[m,e,d]), split(A,C,[m]), split(A,[s],D),!.
past(A,B) :- split(B,A,[r,e,d]), split(A,C,[u,r]),!.
past(A,B) :- split(B,A,[d]), split(A,C,[e]),!.
past(A,B) :- split(B,A,[e,d]),!.
```

The training times for the various systems considered in this research are difficult to compare. Ling does not provide timing results, though we can probably assume based on research comparing symbolic and neural learning algorithms (Shavlik, Mooney, & Towell, 1991) that SPA runs fairly quickly since it is based on C4.5 and that backpropagation took considerably longer. Our tests with FOIL and FOIDL are not directly comparable because they were run on different architectures. The FOIL runs were done on a Sparc 5. For 500 examples, FOIL averaged 48 minutes on the phonetic task with all verbs. The FOIDL experiments ran on a Sparc 2 and averaged 1071 minutes on the same task. Even allowing for the differences in speed of the two machines (about a factor of two), FOIDL is quite a bit slower, probably due largely to the cost of using intentional background and in part to its implementation in Prolog as opposed to C.

## 5. Related Work

### 5.1 Related Work on ILP

Although each of the three features mentioned in the introduction distinguishes FOIDL from most work in Inductive Logic Programming, a number of related pieces of research should be mentioned. The use of intensional background knowledge is the least distinguishing feature since a number of other ILP systems also incorporate this aspect. FOCL (Pazzani & Kibler, 1992), MFOIL (Lavrač & Džeroski, 1994), GRENDEL (Cohen, 1992), FORTE (Richards & Mooney, 1995), and CHILLIN (Zelle & Mooney, 1994a) all use intensional background to some degree in the context of a FOIL-like algorithm. Some other ILP systems which employ intensional background include early ones by Shapiro (1983) and Sammut and Banerji (1986) and more recent ones by Bergadano et al. (1993) and Stahl, Tausend, and Wirth (1993).

The use of implicit negatives is significantly more novel. As described in section 3.2, this approach is considerably different from explicit construction using a closed-world assumption, and therefore can be employed when explicit construction of sufficient negative examples is intractable. Bergadano et al. (1993) allows the user to supply an intensional definition of negative examples that covers a large set of ground instances (e.g `past([a,c,t],X)`, `not(equal(X,[a,c,t,e,d])))`; however, to be equivalent to output completeness, the user would have to explicitly provide a separate intensional negative definition for each positive example. The non-monotonic semantics used to eliminate the need for negative examples in CLAUDIEN (De Raedt & Bruynooghe, 1993) has the same effect as an output completeness assumption in the case where all arguments of the target relation are outputs. However, output completeness permits more flexibility by allowing some arguments to be specified as inputs and only counting as negative examples those extra outputs generated for specific inputs in the training set. FLIP (Bergadano, 1993) provides a method for learning functional programs without negative examples by making an assumption equivalent to output completeness for the functional case. Output completeness is more general in that it permits learning non-functional programs as well. Also, unlike FOIDL, none of these previous methods provide a way of *quantifying* implicit negative coverage in the context of a heuristic top-down specialization algorithm.

The notion of a first-order decision list is unique to FOIDL. The only other ILP system that attempts to learn programs that exploit clause-order and cuts is that of Bergadano et al. (1993). Their paper discusses many problems with learning arbitrary programs with cuts, and the brute-force search used in their approach is intractable for most realistic problems. Instead of addressing the general problem of learning arbitrary programs with cuts, FOIDL is tailored to the specific problem of learning first-order decision lists, which use cuts in a very stylized manner that is particularly useful for functional problems that involve rules with exceptions. Bain and Muggleton (1992) and Bain (1992) discuss a technique which uses negation as failure to handle exceptions. However, using negation as failure is significantly different from decision lists since it simply prevents a clause from covering exceptions rather than learning an additional clause that both over-rides an existing clause *and* specifies the correct output for a set of exceptions.

## 5.2 Related Work on Past-Tense Learning

The shortcomings of most previous work on past-tense learning were reviewed in section 2.2, and the results in section 4 clearly demonstrate the generalization advantage FOIDL exhibits on this problem. However, a couple of issues deserve some additional discussion.

Most of the previous work on this problem has concerned the modelling of various psychological phenomenon, such as the U-shaped learning curve that children exhibit for irregular verbs when acquiring language. This paper has not addressed the issue of psychological validity, rather it has focused on performance accuracy after exposure to a fixed number of training examples. Therefore, we make no specific psychological claims based on our current results.

However, humans can obviously produce the correct past tense of arbitrarily-long novel words, which FOIDL can easily model while fixed-length feature-based representations clearly cannot. Ling also developed a version of SPA that eliminates position dependence and fixed word-length (Ling, 1995) by using a sliding window like that used in NETtalk (Sejnowski & Rosenberg, 1987). A large window is used which includes 15 letters on either side of the current position (padded with blanks if necessary) in order to always include the entire word for all the examples in the corpus. The results on this approach are significantly better than normal SPA but still inferior to FOIDL's results. Also, this approach still requires a fixed-sized input window which prevents it from handling arbitrary-length irregular verbs. Recurrent neural networks could also be used to avoid word-length restrictions (Cotrell & Plunkett, 1991), although it appears that no one has yet applied them to the standard present-tense to past-tense mapping problem. However, we believe the difficulty of training recurrent networks and their relatively poor ability to maintain state information arbitrarily long would limit their performance on this task.

Another issue is that of the comprehensibility and transparency of the learned result. FOIDL's programs for past-tense are short, concise, and very readable; unlike the complicated networks, decision forests, and pure logic programs generated by previous approaches. Ling and Marinov (1993) discusses the possibility of transforming SPA's decision forest into more comprehensible first-order rules; however, the approach of directly learning first-order rules from the data seems clearly preferable.

## 6. Future Work

One obvious topic for future research is FOIDL's cognitive modelling abilities in the context of the past-tense task. Incorporating over-fitting avoidance methods may allow the system to model the U-shaped learning curve in a manner analogous to that demonstrated by Ling and Marinov (1993). Its ability to model human results on generating the past tense of novel psuedo-verbs (e.g., spling → splang) could also be examined and compared to SPA (Ling & Marinov, 1993) and connectionist methods.

Although first-order decision lists represent a fairly general class of programs, currently our only convincing experimental results are on the past-tense problem. Many realistic problems consist of rules with exceptions, and experimental results on additional applications are needed to support the general utility of this representation.

Despite its advantages, the use of intensional background knowledge in ILP incurs a significant performance cost, since examples must be continually reproved when testing alternative literals during specialization. This computation accounts for most of the training time in FOIDL. One approach to improving computational efficiency would be to maintain partial proofs of all examples and incrementally update these proofs as additional literals are added to the clause. This approach would be more like FOIL's approach of maintaining tuples, but would require using a meta-interpreter in Prolog, which incurs its own significant overhead. Efficient use of intensional knowledge in ILP could greatly benefit from work on rapid incremental compilation of logic programs, i.e., incrementally updating compiled code to account for small changes in the definition of a predicate.

FOIDL could potentially benefit from methods for handling noisy data and preventing over-fitting. Pruning methods employed in FOIL and related systems (Quinlan, 1990; Lavrač & Džeroski, 1994) could easily be incorporated. In the decision list framework, an alternative to simply ignoring incorrectly covered examples as noise is to treat them as exceptions to be handled by subsequently learned clauses (as in the uncovering technique discussed in section 3.3).

Theoretical results on the learnability of restricted classes of first-order decision lists is another interesting area for research. Given the results on the PAC-learnability of propositional decision lists (Rivest, 1987) and restricted classes of ILP problems (Džeroski, Muggleton, & Russell, 1992; Cohen, 1994), an appropriately restricted class of first-order decision lists should be PAC-learnable.

## 7. Conclusions

This paper has addressed two main issues: the appropriateness of a first-order learner for the popular past-tense problem, and the problems of previous ILP systems in handling functional tasks whose best representation is rules with exceptions. Our results clearly demonstrate that an ILP system outperforms both the decision-tree and the neural-network systems previously applied to the past-tense task. This is important since there have been very few results showing that a first-order learner performs significantly better than apply-ing propositional learners to the best feature-based encoding of a problem. This research also demonstrates that there is an efficient and effective algorithm for learning concise, comprehensible symbolic programs for a small but interesting subproblem in language acquisition. Finally, our work also shows that it is possible to efficiently learn logic programs which involve cuts and exploit clause order for a particular class of problems, and it demonstrates the usefulness of intensional background and implicit negatives. Solutions to many practical problems seem to require general default rules with characterizable exceptions, and therefore may be best learned using first-order decision lists.

## Acknowledgements

Most of the basic research for this paper was conducted while the first author was on leave at the University of Sydney supported by a grant to Prof. J.R. Quinlan from the Australian Research Council. Thanks to Ross Quinlan for providing this enjoyable and productive opportunity and to both Ross and Mike Cameron-Jones for very important discussions and pointers that greatly aided the development of FOIDL. Thanks also to Ross for aiding us in running the FOIL experiments. Discussions with John Zelle and Cindi Thompson at the University of Texas also influenced this work. Partial support was also provided by grant IRI-9310819 from the National Science Foundation and an MCD fellowship from the University of Texas awarded to the second author.